\documentclass[10pt,twocolumn,letterpaper]{article}

\usepackage{ijcb}
\usepackage{times}
\usepackage{epsfig}
\usepackage{graphicx}
\usepackage{amsmath}
\usepackage{amssymb}
\usepackage{xcolor}
\usepackage{enumitem}
\usepackage{url}
\usepackage[norule,symbol,perpage]{footmisc}

\ijcbfinalcopy 

\ifijcbfinal\fi

\begin{document}

\title{
Modeling Score Distributions and Continuous Covariates: \\
A Bayesian Approach
}

\author{\parbox{16cm}{\centering
    {\large Mel McCurrie$^{1}$,  Hamish Nicholson$^{1,3}$, Walter J. Scheirer$^{1,2}$, and Samuel Anthony$^{1}$}\\
    {\normalsize
    $^1$ Perceptive Automata, Inc.\\
    $^2$ University of Notre Dame\\
    $^3$ Harvard University\\
    }}
}

\maketitle
\thispagestyle{empty}

\begin{abstract}

    Computer Vision practitioners must thoroughly understand their model's performance, but conditional evaluation is complex and error-prone. In biometric verification, model performance over continuous covariates---real-number attributes of images that affect performance---is particularly challenging to study. 
    We develop a generative model of the match and non-match score distributions over continuous covariates and perform inference with modern Bayesian methods. We use mixture models to capture arbitrary distributions and local basis functions to capture non-linear, multivariate trends. 
    Three experiments demonstrate the accuracy and effectiveness of our approach. First, we study the relationship between age and face verification performance and find previous methods may overstate performance and confidence. Second, we study preprocessing for CNNs and find a highly non-linear, multivariate surface of model performance. Our method is accurate and data efficient when evaluated against previous synthetic methods. Third, we demonstrate the novel application of our method to pedestrian tracking and calculate variable thresholds and expected performance while controlling for multiple covariates.

\end{abstract}

    \begin{figure}[t!]
        \centering
        \includegraphics[width=7.5cm]{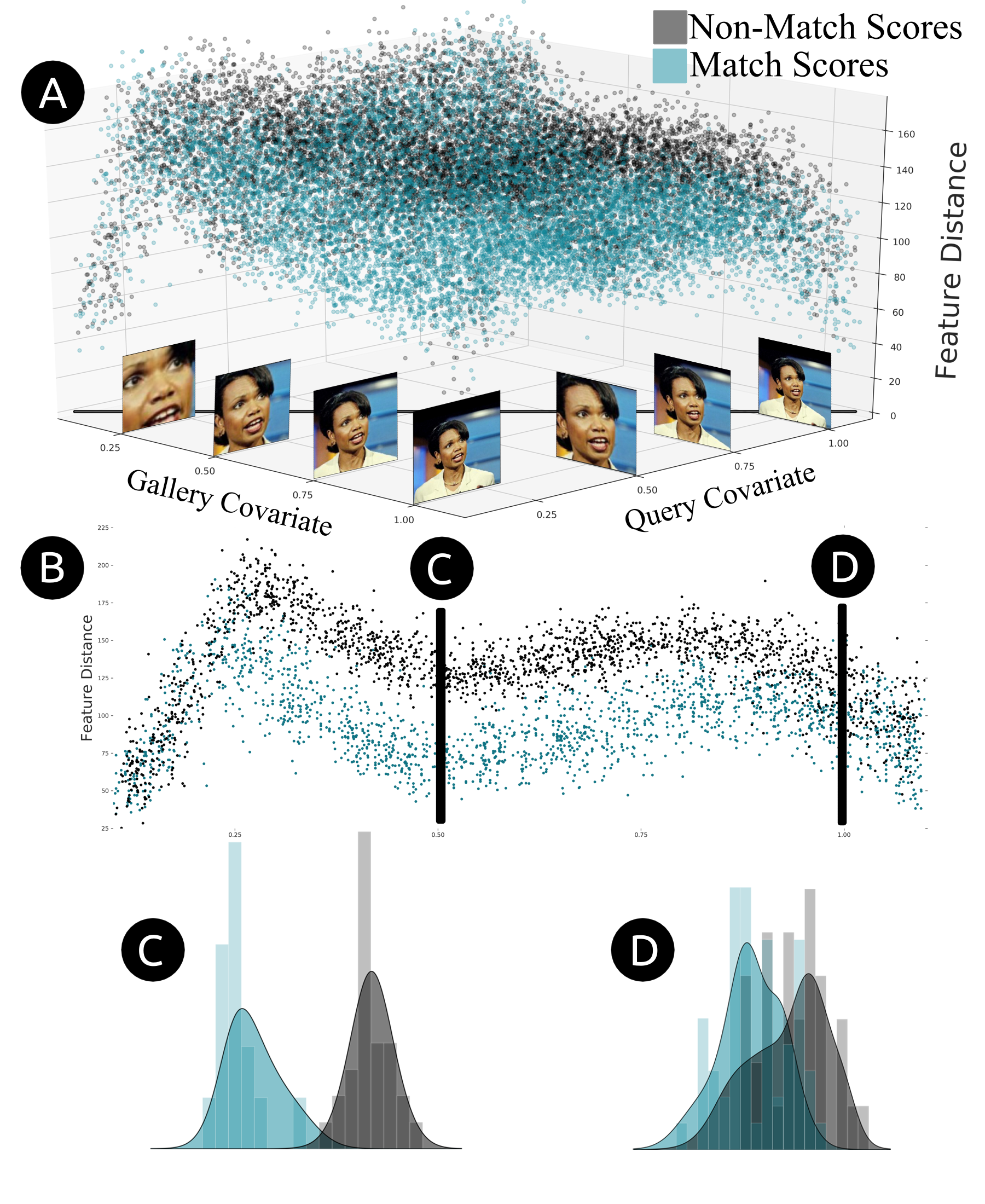}
        \caption[LoF entry]{Our objective is to capture the relationships between multiple continuous covariates and biometric performance. It is impossible to use traditional regression techniques to estimate verification metrics like ROC-curves that are calculated over the entire dataset. Instead, we develop a generative model that allows us to estimate the latent match and non-match distributions and thus metrics at any covariate values. (A) Here we perform pair-wise comparisons between images on the Labeled Faces in the Wild verification dataset where each image is preprocessed with a random scale value, the covariate, before being center cropped and fed into VGGFace2. (B) On the diagonal slice of the multivariate data we can see the match and non-match densities have non-linear trends---a single threshold and an aggregate metric would be inappropriate even for this univariate slice. (C,D) Slices of the data at specific covariate values with few points create noisy empirical densities, but our method estimates the smooth, latent distributions.}
        \label{fig:abs}
    \end{figure}

\section{Introduction}

    Computer Vision practitioners must thoroughly understand their model's performance, but conditional evaluation is complex and error-prone. In biometric verification, model performance over continuous covariates---known, measurable, real-number attributes of images that affect performance---is particularly challenging to study. It is impossible to use traditional regression methods because metrics  like ROC Curves are calculated over an entire dataset, not at individual data points.
    Current methods make strong and simplifying assumptions about the data, commonly treating continuous covariates as discrete or assuming a fixed threshold.  Other studies avoid these statistical complexities and limit their analysis to covariates that can be simulated. These studies generate billions or trillions of data points to perform computationally expensive grid searches over covariates like blur, noise, and occlusion, and still only capture performance at a finite set of points. Especially in the presence of limited data or multiple covariates, current methods for continuous covariates fall short of computer vision practitioners' needs.
    
    Our objective is to capture continuous relationships between multiple continuous covariates and biometric performance without making strong assumptions or synthesizing an unreasonable amount of data.
    Instead of directly modeling the final performance metric, we model the underlying match and non-match distributions of feature distances over continuous covariates with flexible Bayesian methods. To capture arbitrary feature distribution shapes we replace the traditional empirical approach with a mixture model. To capture non-linear trends we model mixture component parameters with local basis functions. To capture uncertainty where data is limited and scale where data is abundant we perform inference with Monte Carlo Sampling and Stochastic Variational Inference on modern hardware. In short we introduce a method for conditional analysis that does the following:
    
    \begin{itemize}[noitemsep]
        \item{Accurately models both the match and non-match score distributions.}
        \item{Captures continuous, non-linear relationships between multiple covariates and model performance.}
        \item{Controls for both Query and Gallery covariates.}
        \item{Generates match and non-match feature distances at arbitrary covariate specifications.}
        \item{Expresses uncertainty where data is limited.}
        \item{Reduces data and compute needs.}
    \end{itemize}
    
    We demonstrate and evaluate our method with three experiments. First, we re-examine a study~\cite{albiero2020does} on the relationship between subject age and verification performance. The study we examine constrains query and gallery pairs to have the same discrete covariate values---we extend this methodology to the continuous case. Ultimately we find that previous studies probably overstate model performance and confidence.
    
    Second, we study the effect of preprocessing on the verification performance of a Convolutional Neural Network. We find extremely low model performance outside of the optimal range of image scale levels, and show how previous works that do not model both query and gallery covariates can understate model robustness. We regenerate the entire Labeled Faces in the Wild~\cite{huang2008labeled} (LFW) dataset at 100 combinations of query and gallery covariate levels to  approximate ground truth model performance at each point. Our method accurately captures the highly non-linear trends in performance when compared to the approximate ground truth. Additionally we apply our method to the ``preferred view", a method to isolate synthetic covariate effects introduced by RichardWebster et al.~\cite{richardwebster2018visual}.
    
    Third, we tackle a previously unstudied problem, and calculate the performance of a pedestrian re-identification model used to track pedestrians over temporal occlusion. We model the match and non-match distribution as functions of elapsed time and control for changes in detection size. We demonstrate how our method can be used during inference to dynamically vary the threshold in a pedestrian tracking scenario to maintain a constant False Positive Rate.

\section{Related Work}
    
    Our work builds upon existing studies that evaluate verification systems under varying conditions. We organize previous work into \textit{Natural Covariate Studies} and \textit{Synthetic Covariate Studies}. Natural covariate studies annotate and analyze attributes like age, gender, and race that are difficult to synthetically alter independent of identity. These studies face a statistical challenge due to limited data. Synthetic covariate studies programmatically alter face attributes like expression and image attributes like blur and noise. These studies are the gold standard, but are expensive to scale and can only be used for some covariates.
    
    \subsection{Natural Covariate Studies}
        Each image in the dataset of a natural covariate study only has one value per covariate of interest. Synthetic techniques cannot be used to generate alternate versions of the dataset. Data is usually distributed unevenly over covariate values resulting in limited data in many regions of covariate space. In these sparsely populated regions conditional performance evaluation methods face a difficult challenge measuring performance with certainty.
     
        The 2002 Face Recognition Vendor Test~\cite{phillips2003face} studied changes in verification and identification performance over changes in covariates including gender, age, and elapsed time. They found different cohorts performed differently, but only considered covariates in the query set, compared identification rates between different dataset configurations, and binned age and elapsed time.
        Mitra et al.~\cite{mitra2007statistical} used a GLM with random effects to predict match rates from covariates like illumination with a normal, linear assumption.
        Scheirer et al.~\cite{scheirer2008predicting} modeled a surface of match score performance with a Support Vector Machine.
        O'Toole et al.~\cite{o2012demographic} studied the effects of race and gender on verification performance. They concentrated on removing race and gender as potential discriminative features by constraining non-match pairs to be of the same race or gender, a technique called ``yoking". 
        Best-Rowden and Jain~\cite{best2017longitudinal} modeled the effects of elapsed time, race, gender, and other covariates on the match score. They fit a multi-level regression model with normal, linear assumptions and estimated confidence intervals by bootstrapping.
        Lu et al.~\cite{lu2019experimental} binned age into seven groups and studied group effects on verification performance.
        Cook et al.~\cite{cook2019demographic} binned age into 2 groups and used a linear regression with bootstrapped confidence intervals to measure effects on performance.
        Most recently Albeiro and Bowyer~\cite{albiero2020does} binned age into three groups, ``young", ``middle", and ``old". They estimated ROC curves for each group, controlling for race and gender, and calculated bootstrapped confidence intervals.

    Previous works that do not model continuous relationships may miss interesting trends in both performance and certainty. Previous works that do model continuous trends concentrate on modeling match scores, usually with the intent of capturing effects.  For the sake of analysis most works make unrealistic assumptions of linear trends or normal feature distributions. In this work we develop a generative model of the continuous match and non-match densities over continuous query and gallery covariates. This allows us to numerically calculate optimal thresholds and expected metrics, as well as empirically estimate performance by simulating specific populations from posterior samples. We do not assume trends are linear, do not assume a specific distribution, and do not assume fixed thresholds. It is worth noting that other fields have developed similar methods for calculating covariate-specific ROC curves, for a thorough review see the work of Rodr{\'\i}guez-{\'A}lvarez et al.~\cite{rodriguez2020rocnreg}.

    \subsection{Synthetic Covariate Studies}
        Synthetic covariate studies reproduce entire datasets and environments under any specified conditions. Computer graphics software can be used to synthesize faces with varying expression~\cite{richardwebster2018visual} and pose~\cite{kortylewski2018empirically} and pedestrians with varying clothing~\cite{pumarola20193dpeople}. Recent developments in generative models enable researchers to manipulate skin tone, hair color, and gender of existing face images~\cite{choi2018stargan} or generate new ones.
        
        Scheirer et al.~\cite{scheirer2014perceptual} simulated varying amounts of occlusion to compare different face detectors. They graph accuracy against the area of face that is visible, calling the graph a psychometric curve. 
        RichardWebster et al.~\cite{richardwebster2018psyphy} used the psychometric curve to display object recognition performance against perturbations like rotation, blur, and contrast. 
        Grm et al.~\cite{grm2017strengths} plotted the mean and standard deviation of face verification accuracy against parameters of perturbations like occlusion, contrast, and compression that were applied to the query set.
        Kortylewski et al.~\cite{kortylewski2018empirically} synthetically rendered face images with different lighting and pose and measured identification performance. They examined joint covariate effects and controlled covariate distributions in their train and test data.
        Nicholson~\cite{nicholson2020psychophysical} manipulated images in the query set of a pedestrian re-identification dataset with blur, noise, compression, and other perturbations, and measured Rank-1 retrieval performance.
        RichardWebster et al.~\cite{richardwebster2018visual} manipulated expression, contrast, blur and other covariates in the query set and measured Rank-1 retrieval performance. They pruned a dataset such that each model's performance is optimal before applying a perturbation, introducing the ``preferred view" of the dataset.
        
        Although progress with generative models and computer graphics is promising, biometric researchers face the unique challenge of manipulating attributes while maintaining a subject's underlying identity. For assessment of attributes that are integral to an identity or at least difficult to vary independently, natural covariate studies remain necessary.  
        Additionally, existing synthetic studies generally limit their analysis to perturbations on the query set, only examining a slice of the true joint metric surface. Even within this slice, perturbations are only applied at finite intervals. Finally, because studies tend to regenerate the entire dataset at every interval, scaling to multiple covariates and increasing the density of the finite intervals would be expensive. Our approach can be used with synthetic techniques to calculate continuous results over joint query and gallery covariates with significantly less data and reduced computational burden.
    
\section{Method}
    
    In this section we describe the general data generation process that underlies our experiments and develop 
    a generative model that captures the densities of the match and non-match distributions over continuous covariates.

    \subsection{Dataset Generation}
    
        A set of $N$ images is collected, and each image is annotated with an identity value and some attributes.  Not all identity values can be unique. During evaluation a pair-wise distance or similarity is calculated between the dataset and itself, resulting in $N^2$ data points. Rows are called ``query" or ``probe" rows and columns called  ``gallery" columns. The complete dataset has $N^2$ data points $\{X_i, y_i\}$ for $i \in \{1,...,N^2\}$ where $y_i$ is the distance or similarity score between two images or their extracted features and $X_i$ is vector of attributes that includes the original query attributes, the original gallery attributes, and user defined interactions between those attributes. One interaction always calculated is a boolean equivalence between query identity and gallery identity that results in the ``match" attribute of $1$ (match) or $0$ (non-match). Usually we are interested in the difference between feature distances whose associated ``match" attribute is $0$ and whose associated ``match" attribute is $1$. We study how this difference varies as a subset of attributes, called ``covariates", varies. We use ``attribute" to describe any latent or known value associated with an image, ``perturbation" to describe an attribute created with synthetic manipulation, ``features" to describe attributes used to calculate feature distances, and ``covariates" to describe the known, measured attributes we study in relation to model performance.
        
        The resulting dataset of attributes and feature distances is partitioned and manipulated based on attributes $\mathbf{X}$. Many works define fixed sets of query and gallery ids, leaving at most $N^{2}/2$ data points. Many works, especially those that calculate metrics derived from the CMC curve, reduce the number of non-match data points by removing all query images with no match in the gallery set. Works in multi-camera pedestrian re-identification reduce the number of match points by removing all data points with equivalent query and gallery camera identity attributes. Most commonly researchers only consider the upper diagonal of the $N$ by $N$ matrix, as symmetric distance functions create a symmetric matrix, and deterministic distance functions cause the diagonal to be $0$ in the case of feature distances, or a maximum value in the case of similarity scores.
        
        In our framework, evaluating models amounts to calculating metrics over the distances $\mathbf{y}$ conditioned on the attributes $\mathbf{X}$.  We treat partitioning and manipulation of a dataset as conditioning on attributes, and thus consider all $N^2$ data points in our analysis. Keeping all match and non-match points increases the data used to estimate the match and non-match distributions. Conditional analysis with feature distances from the matrix diagonal in synthetic experiments produces interesting results. Keeping redundant points from the symmetric matrix provides no additional information but is convenient for modeling over a continuous space and simplifies conditional statements.
            
    \begin{figure}[t!]
        \centering
        \includegraphics[width=8cm]{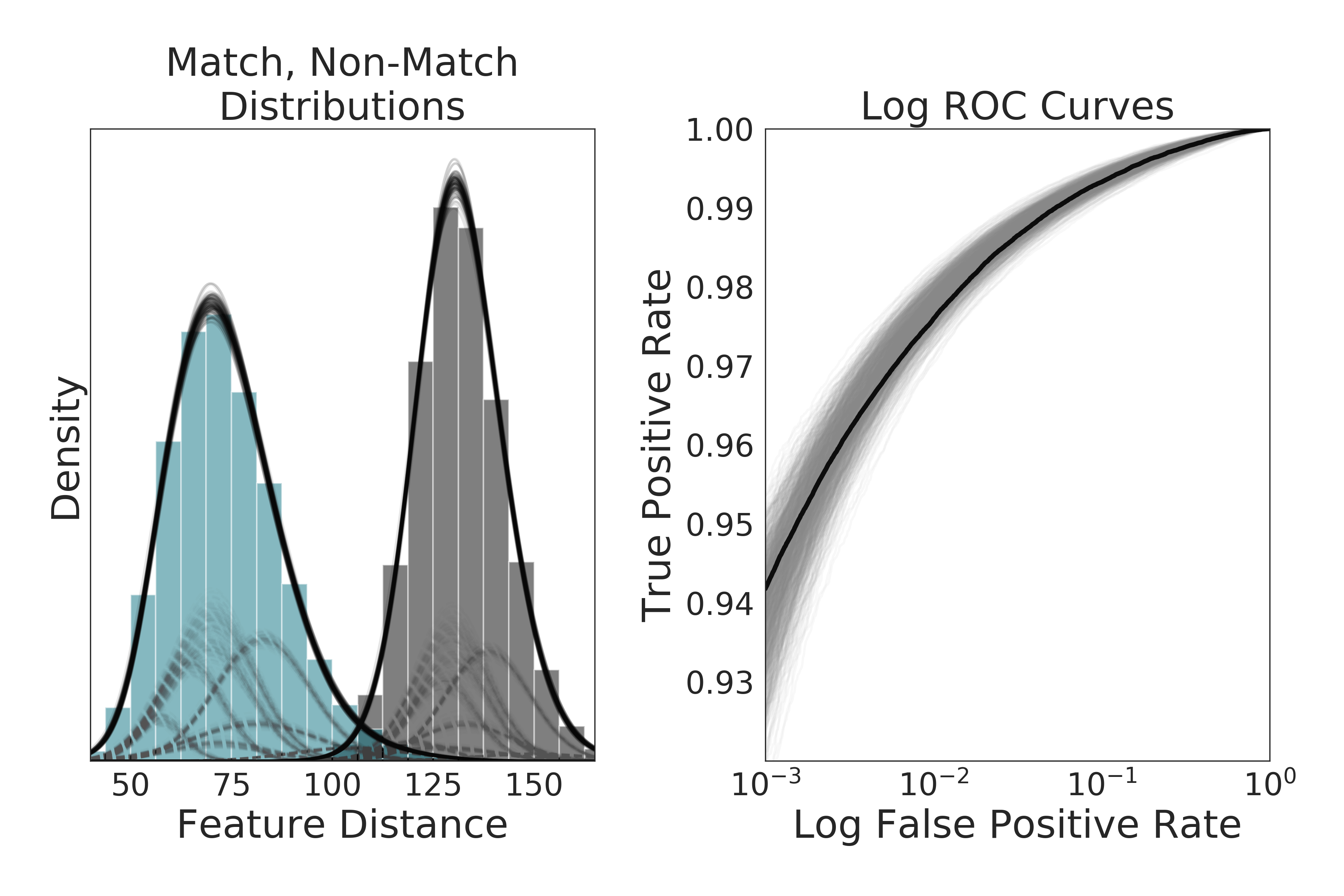}
        \caption[LoF entry]{Estimating match and non-match distributions with mixtures of normals allows us to estimate continuous densities as a function of covariates and capture uncertainty with posterior draws. Here we show this method is accurate on real data even at low False Positive Rates. In this specific example, we use VGGFace2 to extract features from the LFW dataset center cropped at a scale of $0.5$, and calculate euclidean distances between query/gallery pairs. In the left graph we display a teal histogram of match feature distances and a grey histogram of non-match feature distances. Also in the left graph we show posterior draws from the mixture with solid black lines and show individual mixture components scaled by weight with dotted black lines. In the right graph we show a traditional log ROC curve estimated using empirical distributions with a black line, and our method's log ROC curves calculated with posterior draws from the mixtures with grey lines. A traditional ROC curve estimation cannot be extended as a surface over continuous covariates, but our method can.}
        \label{fig:mixture_model}
    \end{figure}

    \subsection{Density Regression}
    
        We introduce a method to estimate the full density of the match and non-match distributions over continuous covariates, allowing us to efficiently estimate model performance with uncertainty at any given range of continuous covariate values with limited data.
        
        The metrics we estimate are derived from the ROC curve, defined as
        
        \begin{equation}\label{eq:roc2}
            \text{TPR}(\mathit{fpr}) = F_M(F_{\bar{M}}^{-1} (\mathit{fpr}))
        \end{equation}

        where $\mathit{fpr} \in [0,1]$ is the false positive rate and $F_M$ and $F_{\bar{M}}$ are the cumulative distribution functions of the match and non-match feature distances, respectively. In addition to ROC curves we summarize performance with the Area Under the ROC Curve (AUC) and True Positive Rate at a fixed false positive rate, usually $10^{-3}$.
        
        It follows from Equation~\ref{eq:roc2} that to estimate an ROC curve we can estimate $F_M$ and $F_{\bar{M}}^{-1}$ independently. Most commonly researchers make few assumptions and use the empirical CDF to estimate $F_M$, use the empirical quantile function to estimate $F_{\bar{M}}^{-1}$, and bootstrap to estimate confidence intervals. This non-parametric, empirical approach is convenient but limiting in the conditional case where we want to estimate
        
        \begin{equation}\label{eq:roc3}
            \text{TPR}(\mathit{fpr}\mid\mathbf{x}) = F_M(F_{\bar{M}}^{-1} (\mathit{fpr}\mid\mathbf{x})\mid\mathbf{x})
        \end{equation}
        
        where $\mathbf{x}$ is our vector of covariates. In continuous space we will have limited or no data at any specific value of $\mathbf{x}$, resulting in uncertain or undefined metrics. We use a more flexible approach and estimate $F_M$ and $F_{\bar{M}}$ with mixtures of normals and numerically invert $F_{\bar{M}}$ to calculate $F_{\bar{M}}^{-1}(fpr)$. Instead of bootstrapping, uncertainty is captured with posterior draws from the mixtures. Figure~\ref{fig:mixture_model} demonstrates this method on real data.
        
        Estimating $F_M$ and $F_{\bar{M}}$ with mixtures of normals allows us to use continuous trends in the data to estimate $\text{TPR}(fpr\mid\mathbf{x})$ at any value of $\mathbf{x}$. This amounts to modeling the weight and location of each component as a function of $\mathbf{x}$, such that a mixture is defined as
        
        \begin{equation}\label{eq:mm}
            \sum_{h=1}^{H} \pi_{h}(\mathbf{x})\mathcal{N}(y_i\mid\mu_{h}(\mathbf{x}), \sigma_{h})
        \end{equation}
        
        where $H$ is the number of components, $\pi_{h}$ models the weights as a function of $\mathbf{x}$, and $\mu_{h}$ models the locations of $\mathbf{x}$.  In order to capture non-linear trends and express uncertainty in regions with limited data we model the locations and weights of our mixtures with local radial basis functions at evenly spaced locations. In some of our experiments we find data is entirely concentrated in small regions of covariate space so we prune basis functions in regions with no data to improve inference.  Additionally, we find both hyperparameter tuning and a Dirichlet Process prior can choose appropriate numbers of components, and opt for hyperparamter tuning in our experiments because it is more consistently stable. For principled estimates of uncertainty we perform MCMC inference using Pyro~\cite{bingham2019pyro} and Pytorch~\cite{paszke2019pytorch}, and scale to large datasets with Stochastic Variational Inference.

\section{Experimental Results}

    In this section we describe the setup and results of three experiments. We study age in face verification, scale in face verification, and temporal occlusion and detection size in pedestrian re-identification.

    \begin{figure}[t!]
        \centering
        \includegraphics[width=8cm]{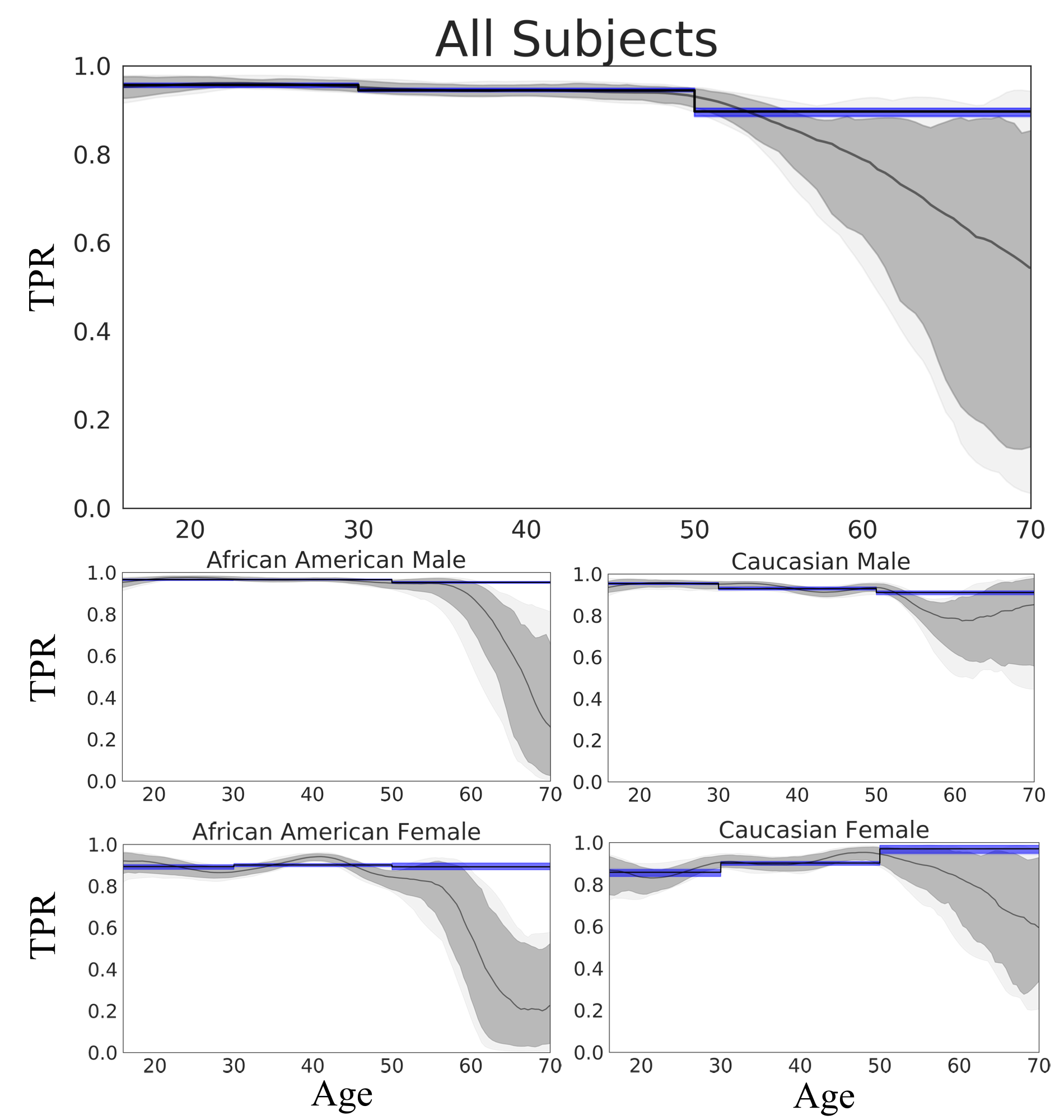}
        \caption[LoF entry]{Verification performance probably decreases with age, but we report very high uncertainty where data is limited and would need more data for conclusive results.  Here, we plot the True Positive Rate at a False Positive Rate of $10^{-3}$ against age. Shown in blue, previous work discretizes age into three bins and estimates the 95\% confidence interval with bootstrapping. We use Bayesian methods to capture continuous trends of performance and uncertainty without binning, and show the 95\% and 99\% credible interval in dark and light grey, respectively. Binning estimates higher performance and higher certainty than our method because it averages over the true continuous trends, effectively weighting results by intra-bin data density. (Note that previous work~\cite{albiero2020does} does not calculate the bottom two graphs, ``African American Female" and ``Caucasian Female", because of limited data. We generated these for illustrative purposes.)}
        \label{fig:morph}
    \end{figure}
    
    \subsection{Does Performance Decrease with Age?}
        Researchers commonly study age when analyzing covariate effects on face verification performance~\cite{lui2009meta}. In this experiment we use the MORPH dataset~\cite{morph} cleaned by the procedure in~\cite{albiero2020does}. This dataset has over 50,000 images that were annotated for age at capture time with additional metadata on gender and race. We compare our method to the recent work of Albeiro et al.~\cite{albiero2020does}.
        
        Albeiro et al. control for age while studying the relationship between age and model performance. Controlling for a covariate amounts to only comparing images that have the same associated covariate value, effectively removing that covariate as a possible discriminative feature. Thus, in experiments with continuous covariates we want to estimate model performance and uncertainty on the diagonal of the query and gallery covariate dimensions where they are exactly equal. Our method uses trends in the feature distances near the diagonal to capture the latent match and non-match distributions on the diagonal, providing continuous estimates of both performance and uncertainty. Previous studies compensate for the lack of data on this diagonal by binning age into discrete groups~\cite{albiero2020does, klare2012face, phillips2003face, cook2019demographic} and calculating intra-bin metrics. Albeiro et al. use ``young" (16-29), ``middle" (30-49), and ``old" (50-70). Binning can misrepresent performance and confidence in three ways. First, it discards the distribution of data within a bin. For example, in the MORPH dataset the quantity of data decreases within the ``old" bin---our intuition says confidence in our metric should decrease too. Second it can average over continuous trends in the data. Even if the old group performs worse, our intuition says performance is not as different between ages 49 and 50 as it is between 30 and 70. Third, it requires a researcher to choose the number and size of bins. Fewer bins results in more confidence but a worse representation of the continuous data, and more bins better captures continuous trends but results in less data per bin and less reasonable confidence estimates from bootstrapping.
    
        In our experiments we use VGGFace2~\cite{cao2018vggface2} to extract features from images in the MORPH dataset and calculate euclidean feature distances. For comparison with our method we bin images into the same age ranges as ~\cite{albiero2020does}, calculate intra-bin true positive rates at a false positive rate of $10^{-3}$, and estimate confidence intervals from one hundred bootstrapped calculations. The results are displayed in blue in Figure~\ref{fig:morph}. For our method we independently normalize the match and non-match feature distances and the query and gallery age and evenly space basis functions for component locations and weights over the normalized space. The entire dataset lies close to the diagonal of query and gallery age so we a priori prune basis functions more than one standard deviation away from any data point. We perform Stochastic Variational Inference in Pyro~\cite{bingham2019pyro} and use 100 samples from our posterior to capture uncertainty. The results are displayed in grey in Figure~\ref{fig:morph}.
        
        Our method shows that performance probably tends to decrease with age. However, our method also expresses very high uncertainty as age increases, reflecting the decrease of data within the old bin. We would need more data in the old group to verify our results. There are two  differences between our results and the binned result in Figure~\ref{fig:morph}. First, our method estimates a lower true positive rate than the ``old" bin. This can be explained by the distribution of data within the old bin. Most of the data tends to come from younger people within the old bin, and thus ``old" bin performance is similar to an average over the true continuous relationship, weighted by data quantity. In fact, we find more, smaller bins capture this relationship. Second, the low certainty from our method contrasts starkly with the high certainty from binned bootstrapping. This can be explained by the large bin size. Within the ``old" bin, from age 50 to age 70, the amount of data decreases. However the ``old" bin captures the total of that data which is enough to cause confident bootstrap results.
    
    \begin{figure}[t!]
        \centering
        \includegraphics[width=8cm]{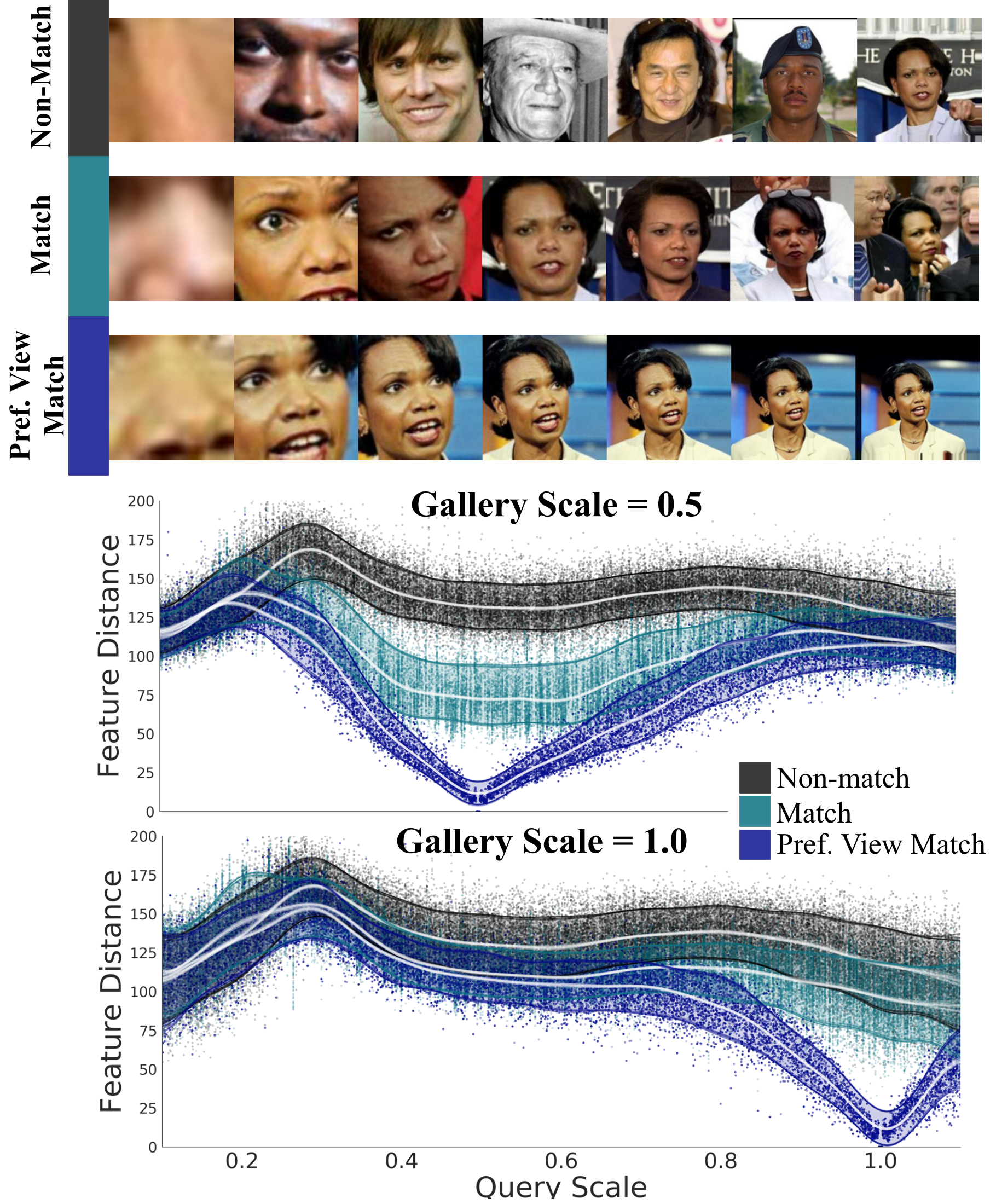}
        \caption[LoF entry]{When the gallery images are scaled by $0.5$ before center cropping, the match and non-match distribution are well-separated for a large range of query scales, indicating robust performance. Works that do not model both query and gallery scales will see a different slice where the gallery scale is $1.0$, and report low robustness. Here, we scatter the non-match, match, and preferred view match feature distances over changes in query scale and show the estimated 80\% densities from our method. The top images make up a legend that exemplifies non-match as comparisons between different identities, match as comparisons between same identities but different original images, and preferred view match as comparisons between same identities with same original images. The closer the preferred view match distribution is to the match distribution, the more the match distribution is explained by the covariate of interest, scale.}
        
        \label{fig:preferredview}
    \end{figure}

    \begin{figure}[t!]
        \centering
        \includegraphics[width=8cm]{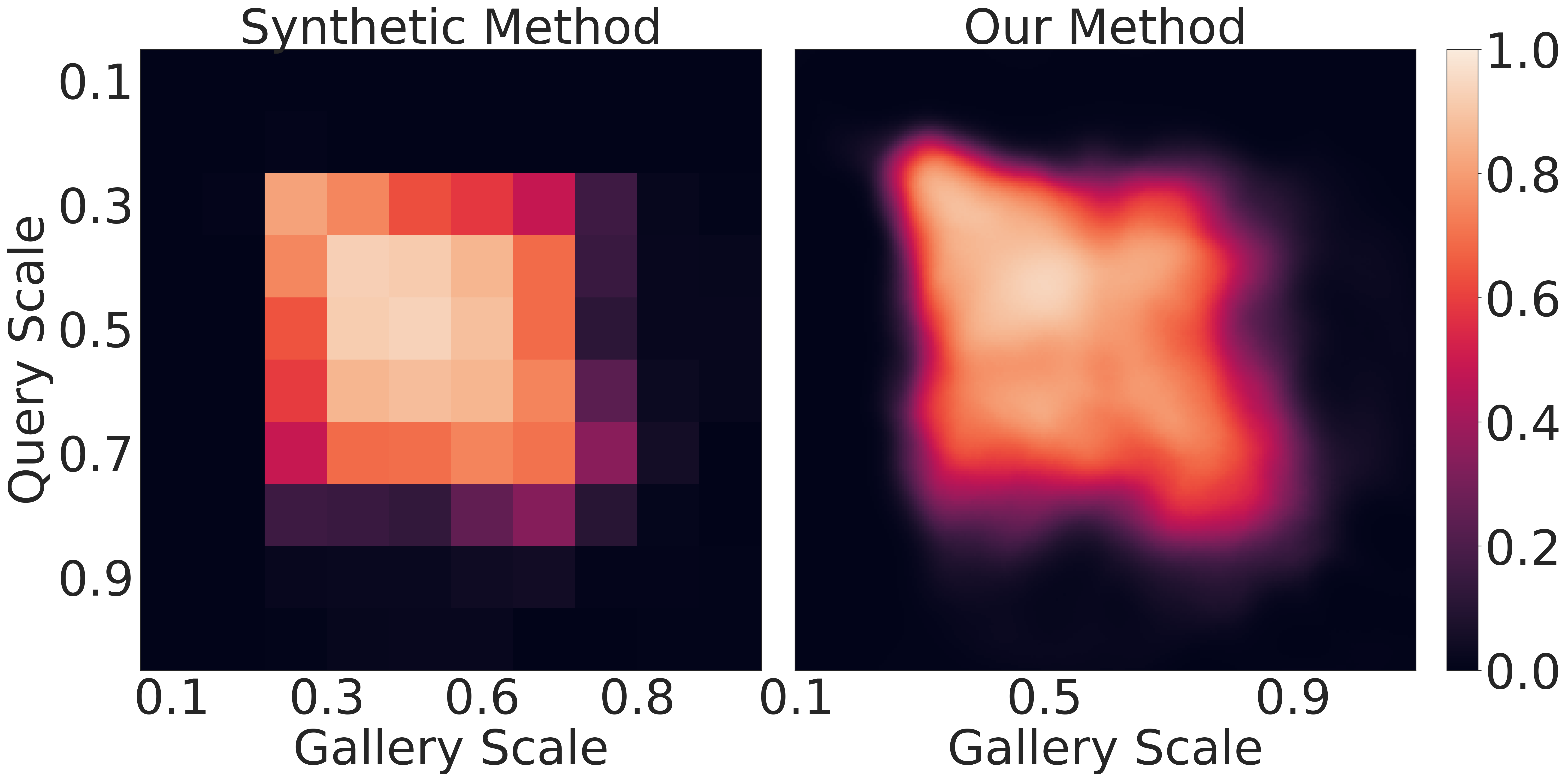}
        \caption[LoF entry]{VGGFace2 performance peaks when LFW query and gallery images are scaled by about $0.5$ before center cropping. Performance is poor if either query or gallery images are scaled less than $0.2$ or greater than $0.8$, and is most robust to changes in scale when query and gallery image scales are the same. Colors indicate True Positive Rate at a False Positive Rate of $10^{-3}$, where 0 is black and 1 is white. An ideal synthetic study, pictured in the left image, captures model performance at a finite number of query and gallery scales. Using over 100 times less data than the synthetic study, our method captures continuous non-linear trends and outputs the dense surface of performance seen in the right image.  Increasing the 10x10 pixel heatmap on the left to be a 250x250 heatmap like the right would require calculating over 10 trillion more feature distances.}
        \label{fig:scale_heatmaps}
    \end{figure}
        
    \subsection{Are Models Robust to Scale Changes?}
        Computer Vision practitioners introduce covariates when they perform image preprocessing for Convolutional Neural Networks. Here we study how the scale parameter in center cropping effects verification performance of the VGGFace2 network. We use the Labeled Faces in the Wild~\cite{huang2008labeled} (LFW) face verification dataset with over 13,000 images. We compare our method to the recent work of RichardWebster et al.~\cite{richardwebster2018visual}.
        
        RichardWebster et al. artificially perturb images to study model performance and robustness, introducing the parameter of a perturbation function as a covariate. At 100 perturbation parameter values, they perturb a 1,000 image subset of the LFW dataset and calculate Rank-1 performance, where the original images are the gallery set and the perturbed images are the query set. They also propose a method to isolate the effects of the covariate of interest. They select a partition of the 1,000 images that maximizes model performance and dataset size using a graph cut algorithm, creating a ``Preferred View" of the dataset. We make several modifications to the original study: we use ROC curve based metrics instead of Rank-1, we use all 13,000 images of the LFW dataset, and we study image scale, which is not one of the perturbations originally studied. Most importantly, we simplify the Preferred View selection algorithm and provide a visual explanation of the Preferred View's importance in Figure~\ref{fig:preferredview}. We simply satisfy the two conditions of the preferred view by selecting feature distances from the diagonal of the symmetric $N$ by $N$ distance matrix. Maximal performance is satisfied because all diagonal feature distances are $0$. Dataset size is maximal because we do not reduce the non-match data points, and theoretically we can use infinite match data points because we can compare any image to itself. We call the match distribution selected from the diagonal the ``preferred view match distribution". The practical benefit of our preferred view formulation is that when we perturb the query and gallery set, feature distances in the preferred view match distribution are no longer $0$, and their positive value is totally explained by the perturbation.
    
        For comparison with our method we pick an evenly spaced ten by ten grid over query and gallery scale dimensions in $[0.1, 1.1]$, where $1.0$ is the original scale of the LFW dataset and $0.1$ is considerably zoomed in. At each point on the grid described by a query scale value $x_q$ and gallery scale value $x_g$ we generate our query set by scaling the entire dataset by $x_q$ and generate our gallery set by scaling the entire dataset by $x_g$. At each point $\{x_g, x_q\}$ we calculate the True Positive Rate at a False Positive Rate of $10^{-3}$ from the feature distances. In total we calculate over 17 billion feature distances. For our method we generate 100 times less data, perturbing each image only once by a random uniform value in $[0.1,1.1]$ to simulate how natural datasets are formed. Based on preliminary data analysis we choose a less smooth prior than previous experiments and reduce the distance between basis functions. Inference is performed the same as our previous experiment. 
        
        Our method shows that VGGFace2 has a very clear optimal range of scale parameters and performance sharply drops outside of that range. There are two major differences between our method and previous methods. First we observe that previous methods only perturb the query images and would misrepresent the robustness and performance of the model. The model performs best where query and gallery scales are equal and zoomed in at a scale of approximately $0.5$. Fixing the gallery set at a scale of $1.0$ and perturbing the query set would show only a poor slice of performance despite robustness on the diagonal. In Figure~\ref{fig:preferredview} we visualize how the match, preferred view match,and non-match distributions change over the query scale at two different fixed gallery scales, demonstrating it is necessary to model both the query and gallery covariates. In Figure~\ref{fig:scale_heatmaps} we model both the query and gallery covariates and visualize the resulting metric surface. Our method produces a dense continuous surface of performance that captures robustness on the diagonal and an optimal scale at $\{0.5, 0.5\}$. In contrast, the synthetic method uses 100 times more data, produces metric results at a finite set of 100 values, and requires exponentially more data and compute to increase the density. Finally, we can consider performance calculated with the 100 simulated datasets to be a gold standard approximation of ground truth model performance and compare our method's estimate at those covariate values to understand its accuracy. From 100 posterior draws our method achieves a high $R^2$ between $0.95$ and $0.98$, the 90\% credible interval.

    \subsection{What is the Correct Threshold for Pedestrian Tracking?}
        Recent improvements in the rapidly growing field of pedestrian re-identification~\cite{leng2019survey} have improved pedestrian tracking performance~\cite{ciaparrone2020deep, wojke2017simple}. Unfortunately evaluation of pedestrian re-identification models has received little attention, and performance specific to tracking models remains unmeasured. In this experiment we demonstrate how our method can be used to estimate a pedestrian re-identification model's performance in a tracking setting. We use the Joint Attention in Autonomous Driving (JAAD) dataset~\cite{rasouli2017they}, a dataset of dashcam videos from a moving vehicle annotated with pedestrian detection, tracking, and various other attributes.
        
        A good pedestrian tracking algorithm tracks pedestrians through temporal occlusion---when objects like other pedestrians and cars obstruct the camera view. It is common to use Convolutional Neural Networks trained for pedestrian re-identification to measure the similarity between two detected pedestrians, and use a hard threshold on feature distances to determine if two pedestrians are the same. However, a hard threshold is unlikely to be optimal given the wide range of conditions in 2D pedestrian tracking. Our intuition tells us that pedestrian appearance probably changes more with elapsed time, and far away pedestrians will have less discriminative features than those closer to the car. In the 2D tracking setting these effects can be captured by elapsed time between detections and bounding box sizes of the two detections.  We examine these three covariates, model the threshold as a function of covariates and a fixed False Positive Rate, and estimate our expected True Positive Rate in specific conditions.
        
        \begin{figure}[t!]
            \centering
            \includegraphics[width=8cm]{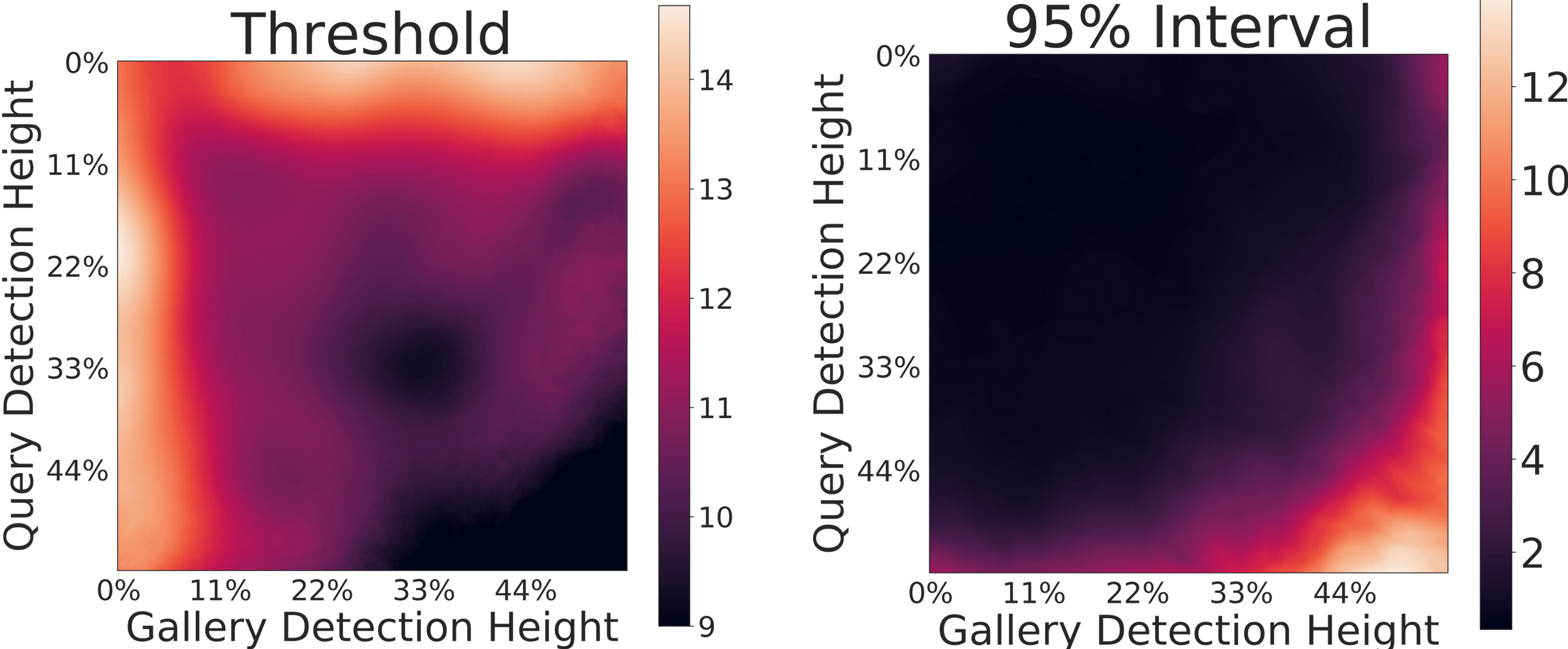}
            \caption[LoF entry]{We show that the threshold needed to maintain a False Positive Rate of $10^{-3}$ is drastically different for different detection sizes. When a small detection is compared to a medium to large detection a very high threshold is required. As the query and gallery detection height each get closer to 33\% of the frame height we can use a much lower threshold. Our confidence for large detections is much lower as there are fewer examples in the dataset.}
            \label{fig:threshold_and_uncertainty}
        \end{figure}

        This task is fundamentally a verification task so we maintain the experimental setup of our previous experiments. Pedestrian images are extracted by cropping detections from the unoccluded subset of the JAAD test. We we extract features with a high performing pedestrian re-identification model, OSNet~\cite{zhou2019omni} and calculate euclidean feature distances. We model the match distribution as a function of query detection height, gallery detection height, and the time between image capture. We model the non-match distribution as a function of query detection height and gallery detection height. We represent detection height as a percentage of the frame height, independently normalize the covariate dimensions, use a smooth prior, and prune our basis functions in regions more than one standard deviation away from any data point. Inference is performed the same as previous experiments. 
        
        We calculate thresholds at a False Positive Rate of $10^{-3}$. Specifically, we numerically invert $F_{\bar{M}}$ to calculate $F_{\bar{M}}^{-1}(10^{-3}\mid\mathbf{x})$ where $\mathbf{x}$ is specific pair of query box height and gallery box height values. We display the results and our confidence in those results as a heatmap in Figure~\ref{fig:scale_heatmaps}. The most obvious trend we find is that smaller detections need a much higher threshold when being compared with medium to larger detections. Our method also captures high uncertainty for large boxes where data is limited. We we would err on the side of caution and choose a high threshold at the edge of our 95\% credible interval when encountering large boxes in practice. 
        
\begin{figure}[t!]
        \centering
        \includegraphics[width=8cm]{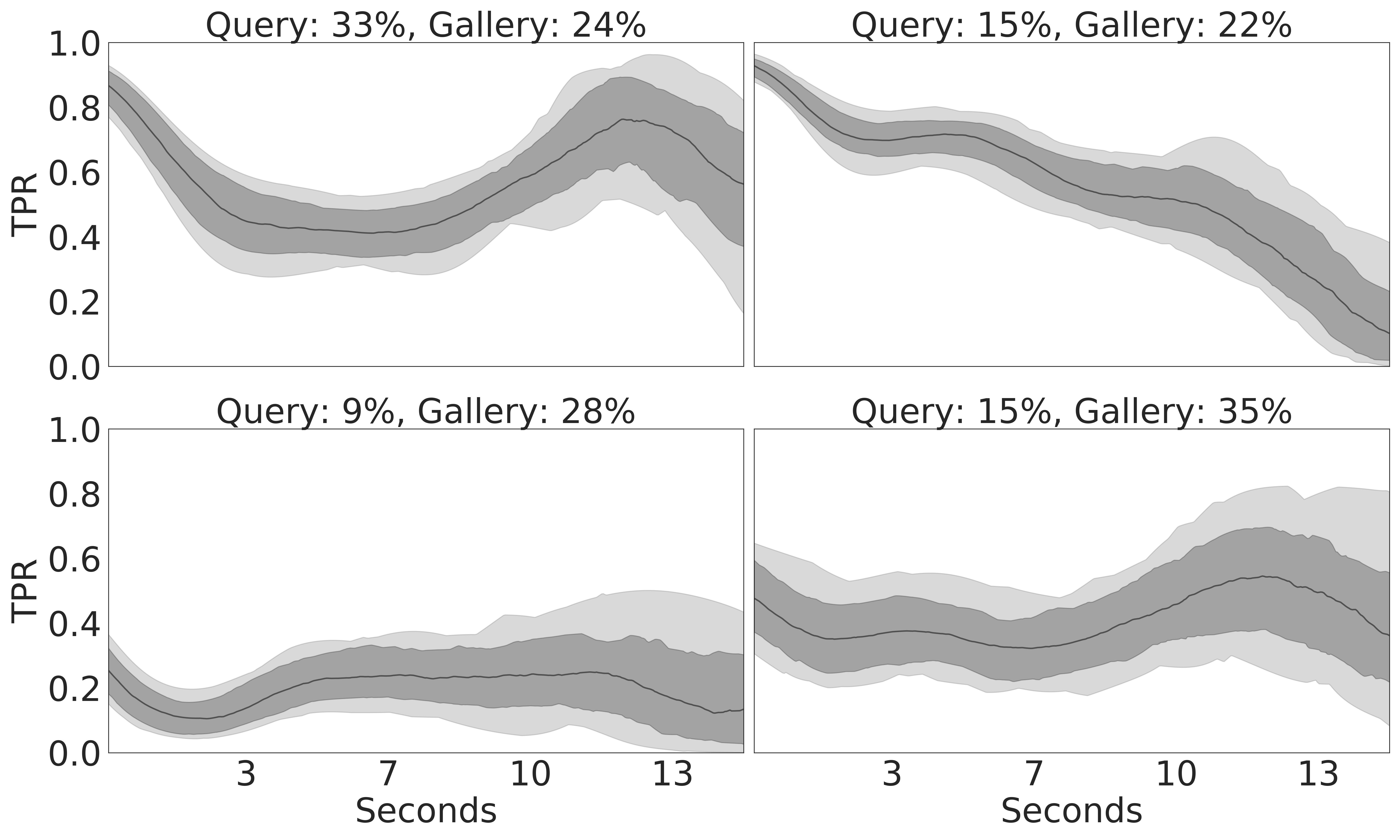}
        \caption[LoF entry]{Even after adjusting thresholds, different query and gallery detection sizes will perform differently over temporal occlusions in a pedestrian tracking scenario. Here we graph our method's 90\% and 98\% credible interval for the estimated True Positive Rate at a False Positive Rate of $10^{-3}$ against the number of seconds between pedestrian detections. Each graph is for a specific combination of query and gallery detection heights, calculated as the percent of the frame height. For example in the top left graph we show that a detection with a height that is 33\% of the frame height will correctly be matched with detections with a height that is 24\% of the frame about 80\% of the time after 12 seconds of occlusion. In general we observe that medium size boxes perform best, and performance is negatively correlated with seconds. However there are some interesting non-linear trends like the late peaks in the top-left and bottom-right graphs.}
        \label{fig:sample_box_sizes}
    \end{figure}

        Modeling the match distribution to estimate the True Positive Rate in a tracking dataset requires an extra covariate. The majority of feature distances in the match distribution will come from images a very short time apart resulting in overstated performance and unreasonably high confidence similar to the binning method in our age experiments. Estimating our match distribution as a function of elapsed time between images allows us to capture decreasing performance and decreasing confidence over increasing temporal occlusions. In Figure~\ref{fig:sample_box_sizes} we visualize these high dimensional results by graphing the expected True Positive Rate against elapsed time for several specific pairs of gallery detection height and query detection height. In general we notice overall performance decreases rapidly within the first few seconds of elapsed time and uncertainty is very high at a larger elapsed time, reflecting the small amount of long tracks in the dataset. Contrary to our intuitions we see a local peak in performance after ten seconds of occlusion at some query/gallery detection height combinations. This is likely caused by by tracks where the car and pedestrian are stationary so pedestrians' features are fairly constant throughout the tracks.

\section{Discussion}

    Throughout our experiments we found our method was best used together with previous works. Binning methods are popular because they are simple, intuitive, and easy to debug.  We found it was useful to use different numbers and sizes of bins to verify our own results and explore the data. Synthetic methods are expensive, but accurate. We used a combination of sparse synthetic trials and our Bayesian method to decide on the final range of image scale values ($[0.1,1.1]$)  studied in our second experiment.  Measuring conditional model performance is a complex task. Claims about model performance require advanced methods and domain knowledge. We present a tool for the computer vision practitioner that makes it easier.
    
{\small
\bibliographystyle{ieee}
\bibliography{main}
}

\end{document}